\documentclass[10pt,twocolumn,twoside]{IEEEtran}

\usepackage{amsmath,amssymb,amsthm,epsfig,dsfont,color,subfigure,empheq,enumerate}
\usepackage{stfloats}
\usepackage{algorithmic,algorithm}
\newtheorem{proposition}{Proposition}
\newtheorem{lemma}{Lemma}

\theoremstyle{remark}

\begin{document}
\bstctlcite{IEEEexample:BSTcontrol}
\title{Distributed Robust Power System State Estimation}

\author{Vassilis Kekatos,~\IEEEmembership{Member,~IEEE,} and Georgios B.
Giannakis*,~\IEEEmembership{Fellow,~IEEE} %
\thanks{The authors are with the ECE Dept., University of Minnesota,
Minneapolis, MN 55455, USA. Dr. Kekatos is also with
the Computer Engnr. \& Informatics Dept., University of Patras, Greece. 
Emails:\{kekatos,georgios\}@umn.edu}}


\maketitle

\begin{abstract}
Deregulation of energy markets, penetration of renewables, advanced metering capabilities, and the urge for situational awareness, all call for system-wide power system state estimation (PSSE). Implementing a centralized estimator though is practically infeasible due to the complexity scale of an interconnection, the communication bottleneck in real-time monitoring, regional disclosure policies, and reliability issues. In this context, distributed PSSE methods are treated here under a unified and systematic framework. A novel algorithm is developed based on the alternating direction method of multipliers. It leverages existing PSSE solvers, respects privacy policies, exhibits low communication load, and its convergence to the centralized estimates is guaranteed even in the absence of local observability. Beyond the conventional least-squares based PSSE, the decentralized framework accommodates a robust state estimator. By exploiting interesting links to the compressive sampling advances, the latter jointly estimates the state and identifies corrupted measurements. The novel algorithms are numerically evaluated using the IEEE 14-, 118-bus, and a 4,200-bus benchmarks. Simulations demonstrate that the attainable accuracy can be reached within a few inter-area exchanges, while largest residual tests are outperformed.
\end{abstract}


\section{Introduction}\label{sec:intro}
Power system state estimation (PSSE) has been traditionally performed at regional control centers with limited interaction. However, due to the deregulation of energy markets, large amounts of power are transferred over high-rate, long-distance lines spanning several control areas \cite{ExpositoPROC11}. These so-called tie lines, originally constructed for emergency situations, are now fully operational and must be accurately monitored. Since any control area can be strongly affected by events and decisions elsewhere, independent system operators (ISOs) can no longer operate in a truly independent fashion. The ongoing penetration of renewable sources further intensifies inter-area power transfers, while their intermittent nature necessitates more frequent state acquisition. At the same time, the advances in metering infrastructure are unprecedented: phasor measurement units (PMUs) provide finely-sampled voltage and current phasors, synchronized across the grid; smart meters reach the distribution level; and networked processors are being installed throughout the grid \cite{ReCeTh10}, \cite{ExpositoPROC11}. The abundance and diversity of measurements offer advanced monitoring capabilities, but processing them constitutes a major challenge, which is exacerbated in the presence of malicious data attacks and bad data \cite{Kosut11}, \cite[Ch.~5-6]{AburExpositoBook}.

There are two key issues in modernizing the power grid monitoring infrastructure: Firstly, PSSE should be performed at the interconnection level. Yet an interconnection may include thousands of buses, while 2-3 measurements per state are typically needed. Requiring also real-time processing along with resilience to corrupted data render centralized state estimation computationally formidable. Further, a centralized approach is vulnerable and is not flexible when it comes to policy and privacy issues. Secondly, decentralizing information processing for the power grid can be performed at several hierarchies \cite{ExpositoPROC11}: PMU measurements can be processed by phasor data concentrators (PDCs) \cite{PhTh08}; conventional supervisory control and data acquisition (SCADA) measurements together with PDC fused data can be aggregated by the ISO; and finally, estimates from ISOs can be merged at the interconnection level. These considerations corroborate that distributed PSSE and bad data analysis are essential for realizing the smart grid vision.

Existing distributed methods for PSSE and bad data analysis are reviewed in Section \ref{sec:review}. The PSSE problem, its unique requirements and challenges are highlighted in Section \ref{sec:problem}. In Section \ref{sec:DPSSE}, a new distributed PSSE methodology is developed. Based on the alternating direction method of multipliers \cite{Boyd10}, a systematic cooperation between local control centers is enabled with unique features: it facilitates several practical PSSE formulations; it lowers the overhead for inter-area information exchanges; its convergence is guaranteed regardless of local observability or parameter tuning; and the resultant algorithm can be executed by solvers already in use at local control centers. Building on this framework, a robust decentralized estimator is derived in Section \ref{sec:DBDA}. Different from the conventional two-step bad data analysis, the novel approach implements Huber's M-estimator \cite{AburExpositoBook} in a decentralized manner, while PSSE is accomplished jointly with bad data removal. Leveraging sparsity of the introduced bad data vectors, the new algorithm augments standard PSSE solvers by a few iterations. The novel robust decentralized algorithms are numerically evaluated in Section \ref{sec:simulations}, and the paper is wrapped up in Section \ref{sec:conclusions}. \textcolor{black}{Regarding notation, lower- (upper-) case boldface letters denote column vectors (matrices), and calligraphic letters stand for sets. The notation $(\cdot)^T$ denotes transposition, while $:=$ defines a symbol variable.}

\section{Related Work}\label{sec:review}
Distributed solutions were pursued since the statistical formulation of PSSE \cite{Schweppe70}, when it was realized that for a chain of serially interconnected areas, Kalman filter-type updates can be invoked incrementally in space \cite[Part III]{Schweppe70}. For arbitrarily connected areas though, a two-level approach with a global coordinator is required \cite{Schweppe70}. Several renditions of this hierarchical approach can be found in \cite{cutsem83}, \cite{Iwamoto89}, \cite{ZhaoAbur05}, \cite{ExpositoPROC11}, \cite{Korres11}. Most of these presume local observability, meaning that local states estimated excluding boundary bus measurements are uniquely identifiable. Such an assumption may not hold due to bad data removal or because PSSE is performed at a lower than the ISO level. The need for a coordinator hinders the system's reliability, while the sought algorithms may be infeasible due to computational, communication, or policy limitations.

Decentralized PSSE solutions include block Jacobi iterations \cite{LinLin94}, \cite{Conejo07}, and an approximate algorithm developed from the optimality conditions involved \cite{Falcao95}. However, these methods assume again local observability and convergence is not always guaranteed. The auxiliary problem principle is used in \cite{Ebrah00}, but several parameters must be tuned. Local observability is waived in \cite{XieChoiKar11}, where each area is envisioned to maintain a copy of the entire high-dimensional state vector. A first-order algorithm is proposed, yet its linear convergence scales unfavorably with the interconnection size. \textcolor{black}{For a review on multi-area PSSE, see also \cite{MASEsurvey}.}

Grossly corrupted SCADA data can potentially deteriorate PSSE results. Hence, these meter readings (a.k.a. bad data) should be identified or at least detected in a measurement set. Statistical tests, such as the $\chi^2$- and the largest normalized residual tests, are typically employed for bad data detection and identification, respectively \cite{Mo00}. Both tests rely on the LS-estimated residuals and can thus run only after PSSE has been completed. Whenever a bad datum has been identified, PSSE must rerun by ignoring that datum. Alternatively, robust estimators, such as the least-absolute deviation, the least median of squares, or Huber's estimator have been considered \cite{CelikAbur92,ElKeib92,Mili94,AburExpositoBook}. Towards a multi-area setup, most existing distributed PSSE methods rely on the two aforementioned tests. Even though metering reliability is improved in the smart grid realm, bad data analysis is still a major concern especially in the face of malicious data attack threats. Stealth attacks in power meter infrastructure are studied in \cite{LiuReiNing09}, \cite{Kosut11}. In the absence of such attacks, $\ell_1$-norm based methods have been developed in \cite{Kosut11}, \cite{XuWaTang11}, and \cite{DuanYangScharf11}.

\section{Problem Formulation and Preliminaries}\label{sec:problem}
Consider an interconnected power system consisting of $K$ control areas, where each area comprises a subset of buses supervised by its own control center. The latter is able to (i) collect the electrical measurements recorded at area buses; (ii) reliably communicate with neighboring control centers; and (iii) carry out minimal computational tasks, such as solving a (non)linear least-squares (LS) problem. \textcolor{black}{As usual, measurements are assumed to be time-synchronized within and across control area footprints.} A control area here is not confined to an ISO region, but it can also model entities residing at lower grid levels, such as a substation or a PDC \cite{PhTh08}. Alternatively, a control area can be the local balancing area under a regional balancing authority's footprint; or under a micro-grid setup, a control area may degenerate even to a single bus.

\textcolor{black}{Suppose that $M_k$ measurements aggregated at the $k$-th area are concatenated in $\tilde{\mathbf{z}}_k\in\mathbb{R}^{M_k}$, and obey the model
\begin{equation}\label{eq:nl_model0}
\tilde{\mathbf{z}}_k=\tilde{h}_k(\mathbf{x}_k) + \tilde{\mathbf{w}}_k
\end{equation}
where $\mathbf{x}_k\in \mathbb{R}^{N_k}$ contains the subset of the interconnected power system states involved in $\tilde{\mathbf{z}}_k$; $\tilde{h}_k$ is a vector of $M_k$ functions; and $\tilde{\mathbf{w}}_k$ denotes a disturbance term capturing measurement error and modeling inaccuracies. Error vectors $\{\tilde{\mathbf{w}}_k\}_{k=1}^K$ are assumed zero mean, having known covariance matrix $\mathbf{\Sigma}_k$, and independent across areas. To simplify the presentation, the model of \eqref{eq:nl_model0} can be premultiplied by $\mathbf{\Sigma}_k^{-1/2}$ to yield
\begin{equation}\label{eq:nl_model}
\mathbf{z}_k=h_k(\mathbf{x}_k) + \mathbf{w}_k
\end{equation}
where $\mathbf{z}_k:=\mathbf{\Sigma}^{-1/2}\tilde{\mathbf{z}}_k$ and similarly for $h(\mathbf{x}_k)$ and $\mathbf{w}_k$. Hence, the noise term in \eqref{eq:nl_model} has identity covariance matrix.}

Functions $\{h_k(\mathbf{x}_k)\}_{k=1}^K$ depend on the system's admittance matrix and are in general non-linear, except when PMUs are involved and complex quantities are expressed in rectangular representations. Performing state estimation with non-linear $h_k$'s entails solving non-convex optimization problems. Typically, such models are iteratively linearized via the Gauss-Newton method, or by resorting to the so called DC approximation \cite{Mo00}, \cite{AburExpositoBook}. Either way, one arrives at the following computationally ubiquitous linear model (cf. \eqref{eq:nl_model})
\begin{equation}\label{eq:model}
\mathbf{z}_k=\mathbf{H}_k\mathbf{x}_k + \mathbf{w}_k
\end{equation}
where $\mathbf{H}_k\in\mathbb{R}^{M_k\times N_k}$ is known (Jacobian). When $\{h_k\}$s are non-linear, \eqref{eq:model} is the linearized model per Gauss-Newton iteration.

\begin{figure}
\centering
\includegraphics[width=0.80\linewidth]{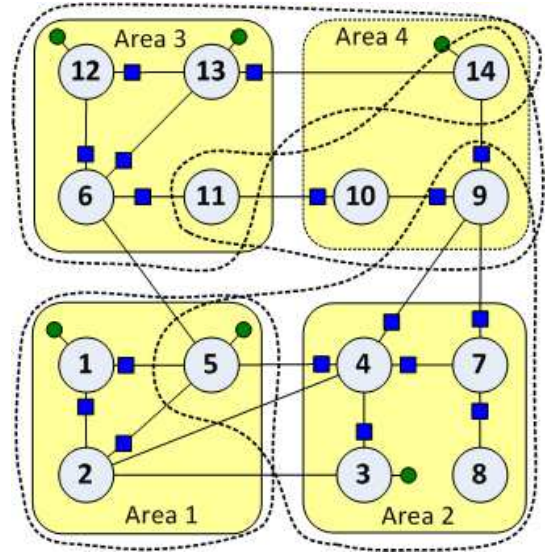}
\caption{The IEEE 14-bus system partitioned into four areas \cite{PSTCA,Korres11}. Dotted lassos show the buses belonging to area state vectors $\mathbf{x}_k$. PMU bus voltage (line current) measurements depicted by green circles (blue squares).}
\label{fig:ieee14}
\end{figure}

PSSE could be performed locally at each area. Specifically, area $k$ could aim at solving
\begin{equation}\label{eq:local_problem}
\min_{\mathbf{x}_k\in\mathcal{X}_k} f_k(\mathbf{x}_k;\mathbf{z}_k,\mathbf{H}_k)
\end{equation}
where $f_k(\cdot)$ is a convex function of $\mathbf{x}_k$ for the model in \eqref{eq:model}; and the convex set $\mathcal{X}_k$ captures possible prior information, such as zero-injection buses, short circuits, or operational limits \cite{Mo00}, \cite{AburExpositoBook}. Typically, $f_k$ is chosen equal to $\frac{1}{2} \|\mathbf{z}_k-\mathbf{H}_k\mathbf{x}_k\|_2^2$. For this choice, the minimizer of \eqref{eq:local_problem} is the LS estimate (LSE), which yields the maximum-likelihood estimate (MLE) of $\mathbf{x}_k$ if $\mathbf{w}_k$ is Gaussian. To derive other MLEs and/or facilitate bad data removal, alternative forms of $f_k$ can be employed; cf.  Section \ref{sec:DBDA}. For notational simplicity, the dependence of $f_k$ on $\mathbf{z}_k$ and $\mathbf{H}_k$ will be henceforth dropped.

\textcolor{black}{One of the unique PSSE characteristics in interconnected areas is that the local state vectors $\{\mathbf{x}_k\}_{k=1}^K$ overlap partially (cf. the toy interconnection of Fig.~\ref{fig:ieee14}). Supposing that both PMU data (bus voltage and line current measurements) and interconnection states (bus voltages) are expressed in rectangular coordinates, the linear model of \eqref{eq:model} is exact. Area 2 supervises buses $\{3,4,7,8\}$. But since it collects current readings on lines $(7,9)$ and $(4,5)$, its state vector $\mathbf{x}_2$ extends to the bus voltages of $\{5,9\}$ as well. Thus, area 2 shares the states of bus 5 (9) with area 1 (4). Notationally, let the $N\times 1$ vector $\mathbf{x}$ collect all the states. For every two neighboring areas, say $k$ and $l$, identify the intersection of their states as $\mathcal{S}_{kl}$. Let also $\mathbf{x}_{k}[l]$ ($\mathbf{x}_{l}[k]$) be the sub-vector of $\mathbf{x}_k$ ($\mathbf{x}_l$) consisting of their overlapping variables ordered as they appear in $\mathbf{x}$. For example, $\mathbf{x}_3[4]=\mathbf{x}_4[3]$ contain the bus voltages of $\{11,14\}$.}

Solving the $K$ problems of the form \eqref{eq:local_problem} in isolation is clearly suboptimal, let alone the fact that control areas may be locally unobservable even if external states and their associated measurements are ignored. Disagreement on boundary bus estimates over critical tie lines is another important limitation of solving \eqref{eq:local_problem} on a per-area basis. On the other hand, upon defining $\mathcal{X}:=\left\{\mathbf{x}:~\mathbf{x}_k\in\mathcal{X}_k~\forall k\right\}$, jointly optimizing
\begin{equation}\label{eq:global_problem}
\min_{\mathbf{x}\in \mathcal{X}} \sum_{k=1}^Kf_k(\mathbf{x})
\end{equation}
at a central control center waives all these concerns and can considerably improve estimation accuracy. Yet this comes at the expense of impractical computational and communication load, increased vulnerability, and disclosure of internal system structure. Targeting the ``sweet spot'' between these two extremes, a decentralized method is proposed next.

\section{Decentralized PSSE}\label{sec:DPSSE}
Tying the local tasks of \eqref{eq:local_problem} into a single optimization problem equivalent to \eqref{eq:global_problem} can be accomplished by
\begin{align}\label{eq:d_problem}
\min_{\{\mathbf{x}_k\in \mathcal{X}_k\}}~&~\sum_{k=1}^K f_k(\mathbf{x}_k)\\
\textrm{s.t.}~&~ \mathbf{x}_{k}[l]=\mathbf{x}_{l}[k],~\forall~l\in\mathcal{N}_k,~\forall k\nonumber
\end{align}
where $\mathcal{N}_k$ is the set of areas sharing states with area $k$.

The constraints of \eqref{eq:d_problem} force neighboring areas to consent on their shared variables, which renders problems \eqref{eq:d_problem} and \eqref{eq:global_problem} equivalent. But the same constraints couple the estimation tasks across areas. To enable a truly decentralized solution, an auxiliary variable denoted by $\mathbf{x}_{kl}\in\mathbb{R}^{|\mathcal{S}_{kl}|}$ is introduced per pair of interacting areas $k,l$. To keep the notation uncluttered, symbols $\mathbf{x}_{kl}$ and $\mathbf{x}_{lk}$ are used interchangeably for the same variable. Then, \eqref{eq:d_problem} can be alternatively expressed as
\begin{align}\label{eq:d_problem2}
\min_{\{\mathbf{x}_k\in\mathcal{X}_k\},\{\mathbf{x}_{kl}\}}~&~\sum_{k=1}^K f_k(\mathbf{x}_k)\\
\textrm{s.t.} ~&~ \mathbf{x}_{k}[l]=\mathbf{x}_{kl},~\textrm{for all}~l\in \mathcal{N}_k,~k=1,\ldots,K.\nonumber
\end{align}

The novelty here is solving \eqref{eq:d_problem2} using the alternating direction method of multipliers (ADMM), a method that has been successfully applied for distributing several optimization problems; see e.g., \cite{Boyd10} for a review. In ADMM, Lagrange multipliers $\mathbf{v}_{k,l}\in \mathbb{R}^{|\mathcal{S}_{kl}|}$ are introduced for each constraint of \eqref{eq:d_problem2}. Observe that $\mathbf{v}_{k,l}$ and $\mathbf{v}_{l,k}$ correspond to the distinct constraints $\mathbf{x}_{k}[l]=\mathbf{x}_{kl}$ and $\mathbf{x}_{l}[k]=\mathbf{x}_{kl}$, respectively. ADMM then exploits the method of multipliers concatenated with an iteration of the Gauss-Seidel algorithm. Specifically for \eqref{eq:d_problem2}, one first defines the augmented Lagrangian function 
\begin{align}\label{eq:lagrangian}
&L\left(\{\mathbf{x}_k\},\{\mathbf{x}_{kl}\};\{\mathbf{v}_{k,l}\}\right):=\\
&\sum_{k=1}^K
\left[f_k(\mathbf{x}_k)+\sum_{l\in\mathcal{N}_k}\left(\mathbf{v}_{k,l}^T(\mathbf{x}_{k}[l]-\mathbf{x}_{kl})+\frac{c}{2}\left\|\mathbf{x}_k[l]{-}\mathbf{x}_{kl}\right\|_2^2\right)\right]\nonumber
\end{align}
where $c>0$ is a predefined constant. Letting $r$ denote the iteration index, ADMM cycles through three steps:
\begin{subequations}\label{eq:Asteps}
\begin{align}
\{\mathbf{x}_k^{t+1}\}&:=\arg\min_{\{\mathbf{x}_k\in\mathcal{X}_k\}}L\left(\{\mathbf{x}_k\},\{\mathbf{x}_{kl}^{t}\};\{\mathbf{v}_{k,l}^{t}\}\right)\label{eq:Asteps1}\\
\{\mathbf{x}_{kl}^{t+1}\}&:=\arg\min_{\{\mathbf{x}_{kl}\}}L\left(\{\mathbf{x}_k^{t+1}\},\{\mathbf{x}_{kl}\};\{\mathbf{v}_{k,l}^{t}\}\right)\label{eq:Asteps2}\\
\mathbf{v}_{k,l}^{t+1}&:=\mathbf{v}_{k,l}^t + c\left(\mathbf{x}_k^{t+1}[l]-\mathbf{x}_{kl}^{t+1}\right)~\textrm{for all}~k,l.\label{eq:Asteps3}
\end{align}
\end{subequations}
At step \eqref{eq:Asteps1}, $\{\mathbf{x}_k\}$s are updated by minimizing the augmented Lagrangian while keeping $\mathbf{x}_{kl}$ and $\mathbf{v}_{k,l}$ fixed to their previous iteration values; $\mathbf{x}_{kl}$ and $\mathbf{v}_{k,l}$ can be initialized to zero. Likewise, $\mathbf{x}_{kl}$ are updated in \eqref{eq:Asteps2}. Finally, \eqref{eq:Asteps3} is a gradient ascent of $L\left(\{\mathbf{x}_k^{t+1}\},\{\mathbf{x}_{kl}^{t+1}\} ;\{\mathbf{v}_{k,l}\}\right)$ with step size $c$.

Inheriting ADMM features, the minimization in \eqref{eq:Asteps1} decouples over control areas. Moreover, by exploiting the problem structure, the iterations \eqref{eq:Asteps} can be greatly simplified as presented next and detailed in the Appendix.

\begin{proposition}\label{pr:iterations}
The steps in \eqref{eq:Asteps} yield the same $\mathbf{x}_k^t$ iterates as the following steps
\begin{subequations}\label{eq:Csteps}
\begin{align}
&\mathbf{x}_k^{t+1}:=\arg\min_{\mathbf{x}_k\in\mathcal{X}_k}f_k(\mathbf{x}_k) {+} \frac{c}{2} \sum_{\substack{i=1\\ \mathcal{N}_k^i\neq\emptyset}}^{N_k} |\mathcal{N}_k^i|\left(x_k(i){-}p_k^t(i)\right)^2,~\forall k\label{eq:Csteps1}\\
&s_k^{t+1}(i):= \frac{1}{|\mathcal{N}_k^i|}\sum_{l\in \mathcal{N}_k^i}x_l^{t+1}[i],~\forall i~\textrm{with}~\mathcal{N}_k^i\neq\emptyset\label{eq:Csteps2}\\
&p_k^{t+1}(i):=p_k^{t}(i) +  s_k^{t+1}(i) - \frac{x_k^{t}(i) + s_k^{t}(i)}{2},
~\forall i, \mathcal{N}_k^i\neq\emptyset\label{eq:Csteps3}
\end{align}
\end{subequations}
where $x_k(i)$ is the $i$-th entry of $\mathbf{x}_k$; the set $\mathcal{N}_k^i$ consists of the areas sharing the variable $x_k(i)$ with area $k$; and $x_l[i]$ denotes the entry of $\mathbf{x}_l$ corresponding to $x_k(i)$ defined for all $l\in\mathcal{N}_k^i$. Regarding initialization, state variables $\mathbf{x}_k$ are set to arbitrary values $\mathbf{x}_k^0$; variables $p_k^0(i)$ are initialized to $(x_k^0(i)+s_k^0(i))/2$; and $s_k^0(i)$ as in \eqref{eq:Csteps2}.
\end{proposition}

\textcolor{black}{The minimization in \eqref{eq:Csteps1} and the simple update of \eqref{eq:Csteps3} are performed at the local centers. The averaging step of \eqref{eq:Csteps2} is accomplished either through a coordinator, or locally too. Either way, the information revealed per area $k$ is \emph{minimal}. No measurements or regression matrices, but only the boundary bus states need to be exchanged, and only between the interested neighboring areas. States can be initialized to the flat profile \cite{Mo00}, or some prior estimate. To implement the LSE given a non-linear measurement model [cf. (2)], the derived algorithm should be used per Gauss-Newton iteration. At each Gauss-Newton iteration, an approximate linear model is updated locally, and the decentralized iterates are initialized using the latest state estimates.}

\textcolor{black}{Supposing $\{f_k(\mathbf{x}_k)\}_{k=1}^K$ are convex, and $\{\mathcal{X}_k\}_{k=1}^K$ are closed convex sets, the cost in \eqref{eq:d_problem2} evaluated at $\{\mathbf{x}_k^t\}$ generated by \eqref{eq:Csteps} converges under mild conditions (typically met in practice) to the optimal value of \eqref{eq:global_problem} \cite[p.~17]{Boyd10}. Hence, when the power system is globally observable, i.e., the centralized LSE is unique, ADMM iterates minimizing the LS cost converge to it. Notwithstanding, observability is not necessary for the method to converge: if the system is unobservable, ADMM iterates converge to one of the multiple LSEs. The equivalence to the centralized LSE ensures that global observability analysis, assumed given in this work, carries over to the decentralized approach too. In addition, ADMM iterates have been shown to be resilient to asynchronous updates and random failures in the inter-area communication links \cite{Zhu09}.}


For notational convenience, define per area $k$ a diagonal matrix $\mathbf{D}_k$ with $(i,i)$-th entry $|\mathcal{N}_k^i|$. Recall that by definition, $|\mathcal{N}_k^i|$ is zero for strictly local states. Also, define the $N_k$-dimensional vector $\mathbf{p}_k^t$ with $i$-th entry the $p^t_k(i)$ of \eqref{eq:Csteps3} when $|\mathcal{N}_k^i|> 0$, and 0 otherwise. Hence, the second term in the cost of \eqref{eq:Csteps1} is expressed as $\frac{c}{2}\|\mathbf{D}_k^{\frac{1}{2}}(\mathbf{x}_k-\mathbf{p}^t_k)\|_2^2$. For the typical case of the unconstrained LSE, the minimizer of \eqref{eq:Csteps1} is clearly given in closed form by
\begin{align}\label{eq:ls}
\hat{\mathbf{x}}^{t+1}_k:=\left(\mathbf{H}_k^T\mathbf{H}_k+c\mathbf{D}_k\right)^{-1}\left(\mathbf{H}_k^T\mathbf{z}_k+c\mathbf{D}_k\mathbf{p}_k^t\right)
\end{align}
which is a simple yet systematic modification of the local LSE. Existing PSSE software can be straightforwardly exploited for finding \eqref{eq:ls} by simply adding $\sqrt{c}\mathbf{D}_k^{1/2}\mathbf{p}_k^t$ as pseudomeasurements with diagonal loading matrix $\sqrt{c}\mathbf{D}_k^{1/2}$. Note that pseudomeasurements are actually added only for the shared states. As empirically observed in Section \ref{sec:simulations}, the procedure terminates after a few tens of iterations.

\section{Decentralized Bad Data Analysis}\label{sec:DBDA}
Time skews, instrument/communication failures, infrequent instrument calibration, reverse wiring, and parameter uncertainty can yield grossly corrupted SCADA measurements. In the cyber-physical smart grid context, bad data are not simply unintentional metering faults, but can also take the form of malicious data injections \cite{SecurityPROC12}. If an intruder can counterfeit some meters so that the attack vector lies in the range space of the PSSE regression matrix, the attack is undetectable and can arbitrarily perturb state estimates \cite{LiuReiNing09}, \cite{Kosut11}. Excluding these naturally termed stealth attacks, this section focuses on bad data identification. After presenting an outlier-aware estimator, interesting connections are established, to efficiently implement it using the decentralized approach of Section \ref{sec:DPSSE}.

\subsection{Interconnection-Wide Bad Data Identification}\label{subsec:DBDA:global}
The local quantities $\{\mathbf{z}_k,\mathbf{H}_k,\mathbf{w}_k\}_{k=1}^K$ in \eqref{eq:model} can be vertically stacked in $\mathbf{z}$, $\mathbf{H}$, and $\mathbf{w}$, respectively. The interconnection-level model then reads $\mathbf{z}= \mathbf{H}\mathbf{x} +\mathbf{w}$, where the dimensions of $\mathbf{z}$ and $\mathbf{x}$ are $M=\sum_{k=1}^K M_k$ and $N$, in the order given. However, when bad data are present, a more pertinent model is
\begin{equation}\label{eq:model+o}
\mathbf{z}=\mathbf{H}\mathbf{x} + \mathbf{o}+\mathbf{w}
\end{equation}
where $\mathbf{o}$ is an unknown vector with its entry $o(i)$ being non-zero only if $z(i)$ is a bad datum \cite{VKGG11}, \cite{Kosut11}, \cite{DuanYangScharf11}. Recovering both $\mathbf{x}$ and $\mathbf{o}$ essentially reveals the state and identifies faulty measurements. Such a mission however seems rather impossible, since the model in \eqref{eq:model+o} is unobservable even if $\mathbf{H}$ is full column rank. By capitalizing on the sparsity of $\mathbf{o}$ though, interesting results can be obtained \cite{VKGG11}. If $\tau_0$ bad data are expected, then one would ideally wish to solve
\begin{align}\label{eq:ell_0}
\min_{\mathbf{x}\in\mathcal{X},\mathbf{o}}\left\{ \frac{1}{2} \left\|\mathbf{z}-\mathbf{H}\mathbf{x}-\mathbf{o}\right\|_2^2:~\|\mathbf{o}\|_0\leq \tau_0\right\}.
\end{align}
But the $\ell_0$-pseudonorm, i.e., the number of non-zero $o(i)$'s, renders \eqref{eq:ell_0} NP-hard. The problem is computationally intractable even for moderate-sized interconnections and small $\tau_0$. Building on the premise of compressed sensing, a practical robust estimator can be derived after relaxing the $\ell_0$-pseudonorm by the convex $\ell_1$-norm as (see also \cite{VKGG11})
\begin{align}\label{eq:ell_1_c}
\min_{\mathbf{x}\in\mathcal{X},\mathbf{o}}\left\{ \frac{1}{2} \left\|\mathbf{z}-\mathbf{H}\mathbf{x}-\mathbf{o}\right\|_2^2:~\|\mathbf{o}\|_1\leq \tau_1\right\}
\end{align}
for a selected constant $\tau_1>0$, or in the Lagrangian form
\begin{align}\label{eq:ell_1}
(\hat{\mathbf{x}},\hat{\mathbf{o}})\in\arg\min_{\mathbf{x}\in\mathcal{X},\mathbf{o}}~ \frac{1}{2} \left\|\mathbf{z}-\mathbf{H}\mathbf{x}-\mathbf{o}\right\|_2^2 + \lambda\|\mathbf{o}\|_1
\end{align}
where $\lambda$ denotes a positive parameter. The optimization problem in \eqref{eq:ell_1} is a convex quadratic program and can be solved by interior point-based methods. The estimator of \eqref{eq:ell_1} allows for joint state estimation and bad data identification. Even when some measurements are deemed as corrupted, their effect has been already suppressed, and the state estimate remains valid. 

\subsection{Interesting Links}\label{subsec:DBDA:connections}
The two statistical tests traditionally used for bad data analysis rely on the model $\mathbf{z}=\mathbf{H}\mathbf{x}+\mathbf{w}$, and the residual error achieved by the unconstrained LSE. The latter can be expressed as $\mathbf{r}:=\mathbf{P}\mathbf{z}=\mathbf{P}\mathbf{w}$, where $\mathbf{P}:=\mathbf{I} - \mathbf{H}(\mathbf{H}^T\mathbf{H})^{-1}\mathbf{H}^T$ is the so called ``residual sensitivity matrix'' satisfying $\mathbf{P}=\mathbf{P}^2$ \cite[Ch.~5]{AburExpositoBook}. Clearly, when $\mathbf{w}$ is Gaussian, $\mathbf{r}$ is Gaussian too with covariance matrix $\mathbf{P}$. The $\chi^2$-test compares $\|\mathbf{r}\|_2^2$ against a threshold to detect the presence of bad data \cite{Mo00}, \cite{AburExpositoBook}. The largest normalized residual (LNR) test computes
\begin{align}\label{eq:LNR}
\bar{r}_{\max}:=\max_{i\in\{1,\ldots,M\}} \frac{|r(i)|}{\sqrt{P(i,i)}}
\end{align}
where $P(i,i)$ is the $(i,i)$-th entry of $\mathbf{P}$; note that in the absence of critical measurements, $0<P(i,i)\leq 1$ for all $i$. Metric $\bar{r}_{\max}$ is then compared to a prescribed threshold to identify a single bad datum \cite[Sec.~5.7]{AburExpositoBook}. Adopting the proof in \cite[Prop.~1]{Kosut11}, the following claim can be established.

\begin{proposition}\label{pr:l0_lnrt}
The optimization problem in \eqref{eq:ell_0} with $\tau_0=1$ over $\mathcal{X}=\mathbb{R}^N$, and the LNR test of \eqref{eq:LNR} are equivalent in terms of the measurement being detected as bad.
\end{proposition}
For multiple bad data, this connection is unclear. If a measurement is deemed as outlier, PSSE is repeated after discarding this bad datum, the LNR test is re-applied, and the process iterates till no corrupted data are identified. Even though rank-one updates can be used to speed up computations, the process becomes complicated for multi-area grids.

Returning to the convex relaxation \eqref{eq:ell_1} for $\mathcal{X}=\mathbb{R}^N$, note that when $\lambda\rightarrow \infty$, the minimizer $\hat{\mathbf{o}}$ becomes zero, and thus, $\hat{\mathbf{x}}$ reduces to the LSE. On the contrary, by letting $\lambda \rightarrow 0^{+}$ \cite{LiSwe98}, the solution $\hat{\mathbf{x}}$ coincides with $\arg\min_{\mathbf{x}}\|\mathbf{z}-\mathbf{H}\mathbf{x}\|_1$, meaning the least-absolute value estimator \cite{CelikAbur92}, \cite{ElKeib92}.

For finite $\lambda>0$, $\hat{\mathbf{x}}$ of \eqref{eq:ell_1} corresponds to Huber's M-estimator; see \cite{VKGG11} and references therein. Based on this connection and assuming Gaussian $\mathbf{w}$, tuning parameter $\lambda$ can be set to 1.34, which makes the estimator 95\% asymptotically efficient for bad data-free measurements \cite[p.~26]{MaMaYo06}.

Alternatively, Huber's estimate can be expressed as the $\mathbf{x}$-minimizer of $\min_{\mathbf{x}\in\mathcal{X},\boldsymbol{\omega}} \frac{1}{2}\|\boldsymbol{\omega}\|_2^2 + \lambda \|\mathbf{z}-\mathbf{H}\mathbf{x}-\boldsymbol{\omega}\|_1$ as shown in \cite{MaMu00}. The bad data identification performance of this minimization is analyzed in \cite{XuWaTang11}.

To solve \eqref{eq:ell_1}, one can first minimize over $\mathbf{x}$ and then over $\mathbf{o}$. For $\mathcal{X}=\mathbb{R}^N$, the $\mathbf{x}$ minimizing \eqref{eq:ell_1} is $(\mathbf{H}^T\mathbf{H})^{-1}\mathbf{H}^T (\mathbf{z}-\mathbf{o})$, and thus, minimizing \eqref{eq:ell_1} reduces to \cite{Kosut11}, \cite{DuanYangScharf11}
\begin{align}\label{eq:ell_1_o}
\min_{\mathbf{o}}~ \frac{1}{2} \left\|\mathbf{P}(\mathbf{z}-\mathbf{o})\right\|_2^2 + \lambda\|\mathbf{o}\|_1.
\end{align}
A minimization similar to \eqref{eq:ell_1_o} is derived in \cite{Kosut11} using a generalized likelihood ratio test. By assuming a Bayesian prior $\mathbf{x}\sim \mathcal{N}(\mathbf{0},\mathbf{\Sigma}_x)$, \cite{Kosut11} suggests solving \eqref{eq:ell_1_o}, but with $\mathbf{P}$ substituted by $\mathbf{I}-\mathbf{H}(\mathbf{H}^T\mathbf{H}+\mathbf{\Sigma}_x^{-1})^{-1}\mathbf{H}^T$. In any case, matrix $\mathbf{P}$ couples the minimization over $\mathbf{o}$ across areas and complicates a decentralized implementation. An efficient distributed algorithm for solving \eqref{eq:ell_1} is developed next.

\color{black}

\subsection{Distributed Robust Algorithm}\label{subsec:DBDA:decentralized}
Consider now the system-wide minimization in \eqref{eq:ell_1} under the decentralized PSSE format of Section \ref{sec:DPSSE}. To this end, partition $\mathbf{o}$ into subvectors $\mathbf{o}_k$'s conforming to the partition of $\mathbf{z}$ into $\mathbf{z}_k$'s, and define the local functions
\begin{align}\label{eq:huber}
f_k(\mathbf{x}_k,\mathbf{o}_k):= \frac{1}{2} \left\|\mathbf{z}_k-\mathbf{H}_k\mathbf{x}_k-\mathbf{o}_k\right\|_2^2 + \lambda\|\mathbf{o}_k\|_1.
\end{align}
Critically, notice that $\{\mathbf{o}_k\}$ belong to a single area and there is no need for sharing them. Similar to the way \eqref{eq:d_problem2} was obtained from \eqref{eq:global_problem}, the decentralized equivalent of \eqref{eq:ell_1} is
\begin{align}\label{eq:d_ell_1}
&\min_{\{\mathbf{x}_k\in\mathcal{X}_k\},\{\mathbf{o}_k\},\{\mathbf{x}_{kl}\}}~~\sum_{k=1}^K f_k(\mathbf{x}_k,\mathbf{o}_k)\\
&\quad\quad\textrm{s.t.} ~~ \mathbf{x}_{k}[l]=\mathbf{x}_{kl},~\textrm{for all}~l\in \mathcal{N}_k,~k=1,\ldots,K.\nonumber
\end{align}

Proceeding with the ADMM methodology, the augmented Lagrangian of \eqref{eq:d_ell_1} is similar to the one in \eqref{eq:lagrangian} apart from $f_k(\mathbf{x}_k)$ being replaced by $f_k(\mathbf{x}_k,\mathbf{o}_k)$. Having three instead of two primal variable sets offers two algorithmic alternatives: the additional $\mathbf{o}_k$ variables can be jointly optimized either with $\{\mathbf{x}_k\}$s in step \eqref{eq:Asteps1}, or with $\{\mathbf{x}_{kl}\}$s in step \eqref{eq:Asteps2}. The second choice yields computational efficiency as described next.

Two are the key observations here. First, that the augmented Lagrangian of the problem is separable with respect to $\{\mathbf{o}_k\}$ and $\{\mathbf{x}_{kl}\}$, and hence, the optimization required at the ADMM step \eqref{eq:Asteps2} decouples over the two variable sets. Second, the updated $\{\mathbf{o}_k^{t+1}\}$ vectors appear only in $f_k(\mathbf{x}_k,\mathbf{o}_k^{t})$, and do not affect the $\mathbf{x}_{kl}^{t+1}$ and $\mathbf{v}_{k,l}^{t+1}$ updates. Since the iterations of \eqref{eq:Csteps} were derived from \eqref{eq:Asteps} for generic $\{f_k(\mathbf{x}_k)\}$, they can be readily extended to the robust PSSE case as follows.

In step \eqref{eq:Csteps1}, $\{f_k(\mathbf{x}_k)\}$ should be replaced by $\{f_k(\mathbf{x}_k,\mathbf{o}_k^{t})\}$. For $\mathcal{X}_k=\mathbb{R}^{N_k}$ and using the notation introduced before \eqref{eq:ls}, state variables can be updated in closed form as 
\begin{align}\label{eq:x_update}
\mathbf{x}_k^{t+1}:=\left(\mathbf{H}_k^T\mathbf{H}_k+c\mathbf{D}_k\right)^{-1}\left(\mathbf{H}_k^T(\mathbf{z}_k{-}\mathbf{o}_k^{t})+c\mathbf{D}_k\mathbf{p}_k^t\right).
\end{align}
Interestingly, the update of \eqref{eq:x_update} is the LSE of \eqref{eq:ls} slightly modified: measurements $\mathbf{z}_k$ have been substituted by the bad data-compensated measurements $(\mathbf{z}_k-\mathbf{o}_k^t)$.

Steps \eqref{eq:Csteps2}-\eqref{eq:Csteps3} are left untouched; guaranteeing the consent between shared states is unrelated to the local functions $f_k(\mathbf{x}_k)$. The variables $\{\mathbf{o}_k\}$ can be finally updated as the minimizers of $f_k(\mathbf{x}_k^{t+1},\mathbf{o}_k)$ as required by the ADMM step \eqref{eq:Asteps2}. Interestingly, the solution of the latter minimization is provided in closed form too as (cf. \cite[p.~32]{Boyd10})
\begin{align}\label{eq:o_update}
\mathbf{o}_k^{t+1}:=\left[\mathbf{z}_k-\mathbf{H}_k\mathbf{x}_k^{t+1}\right]_{\lambda}^{+}
\end{align}
where $[x]_{\lambda}^{+}$ denotes the simple thresholding operator
\begin{align}\label{eq:thresholding}
[x]_{\lambda}^{+}:=\left\{\begin{array}{ll}
x+\lambda, & x< -\lambda\\
0, & |x|\leq\lambda\\
x-\lambda, & x> \lambda
\end{array}\right.
\end{align}
understood entry-wise in \eqref{eq:o_update}. Intuitively, for measurements with a small tentative absolute residual, the corresponding $o_k(i)$ becomes zero. Otherwise, the bad datum residual is artificially shrunk by $\lambda$ towards zero via a non-zero $o_k(i)$.

\begin{algorithm}[t]
\caption{Decentralized Robust PSSE (D-RPSSE)} \label{alg:DRPSSE}
\begin{algorithmic}[1]
\REQUIRE $\{\mathbf{H}_k,\mathbf{D}_k,\mathbf{z}_k\}$, and positive $c$, $\lambda$.
\STATE Initialize $\{\mathbf{x}_k^0,\mathbf{p}_k^0,\mathbf{s}_k^0\}$ as in \eqref{eq:Csteps}; and $\{\mathbf{o}_k^0\}$ to zero.
\FOR{$t=1,2,\ldots$} 
\STATE Every area updates its $\mathbf{x}_k^{t+1}$ as in \eqref{eq:x_update}.
\STATE Neighboring areas exchange shared state variables.
\STATE Every area updates its $\mathbf{s}_k^{t+1}$ via \eqref{eq:Csteps2}.
\STATE Every area updates its $\mathbf{p}_k^{t+1}$ via \eqref{eq:Csteps3}.
\STATE Every area updates its $\mathbf{o}_k^{t+1}$ through \eqref{eq:o_update}-\eqref{eq:thresholding}.
\ENDFOR
\end{algorithmic}
\end{algorithm}

The novel robust decentralized algorithm, called D-RPSSE hereafter, is tabulated as Alg.~\ref{alg:DRPSSE}. Compared to the decentralized LSE of \eqref{eq:Csteps}-\eqref{eq:ls}, D-RPSSE maintains software compatibility too. On top of adding $\sqrt{c}\mathbf{D}_k^{1/2}\mathbf{p}_k^t$ as pseudo-measurements, resilience to bad data comes by simply off-setting local measurements by $\mathbf{o}_k^{t}$ and via the thresholding rule of \eqref{eq:thresholding}. Robust state estimates and bad data identification are jointly acquired without repeating PSSE as required by the LNRT.

\color{black}

\section{Simulated Tests}\label{sec:simulations}
The developed decentralized state estimators are numerically tested on an Intel Duo Core @ 2.2 GHz (4GB RAM) computer using MATLAB. Two power network benchmarks are initially considered, namely the IEEE 14- and the 118-bus systems; while later a 4,200-bus is generated based on the IEEE 14- and 300-bus systems \cite{PSTCA}. Their admittance matrices and the underlying power system states are obtained using MATPOWER \cite{MATPOWER}. For all systems, the state vector contains the real and imaginary parts of all bus voltages. Measurements consist of PMU recordings on bus voltages and line currents, expressed in rectangular coordinates too. Measurement noise is simulated as independent zero-mean Gaussian with standard deviation per real component $0.01$ and $0.02$ for voltages and currents, respectively \cite{MATPOWER}.

For the IEEE 14-bus grid, PMU sites and types are shown in Fig.~\ref{fig:ieee14}: 6 bus voltage and 17 line current meters yield a total of 46 measurements corresponding to a redundancy ratio of 1.6 \cite{Mo00}. For the IEEE 118-bus grid, PMU sites are selected uniformly at random: 77 bus voltage and 205 line current meters are utilized, yielding a redundancy ratio of 2.4. The IEEE 14-bus grid is partitioned into the 4 areas depicted in Fig.~\ref{fig:ieee14}, while the IEEE 118-bus system is split into 3 areas as in \cite[Fig.~4]{MinAbur05}.

\begin{table}
\renewcommand{\arraystretch}{1.2}
\caption{Average standard deviation per state}
\vspace*{-0.5em}
\centering
\begin{tabular}{|l|r|r|}
\hline
\textbf{Estimator} & IEEE 14-bus grid & IEEE 118-bus grid \\
\hline
Internal LSE			& $3.4\cdot 10^{-3}$	& $4.1\cdot 10^{-4}$ \\
Local LSE				& $3.1\cdot 10^{-3}$	& $4.0\cdot 10^{-4}$	\\
Global LSE			& $1.0\cdot 10^{-3}$	& $2.2\cdot 10^{-4}$	\\
\hline
\end{tabular}
\label{tbl:stds}
\end{table}

A reasonable question is whether interconnection-wide PSSE offers any improvement over local PSSE. To this end, three estimators are numerically compared: First, an estimator that uses only the measurements related to its own area, henceforth called ``internal.'' Second, a ``local'' estimator which extends its state to boundary buses that can be reached via tie line measurements. Lastly, the interconnection-wide or ``global'' estimator. The average standard deviation per state is empirically computed over 100 Monte Carlo runs. Table~\ref{tbl:stds} lists the standard deviations for the two power grids. The IEEE 118-bus grid attains improved estimation accuracy due to its increased redundancy ratio. More importantly, the improvement of the local over the internal estimator is marginal, whereas the accuracy of the global estimator roughly doubles. This observation speaks for the importance of interconnection-wide PSSE even when local observability is guaranteed.

\begin{figure}
\centering
\includegraphics[width=0.82\linewidth]{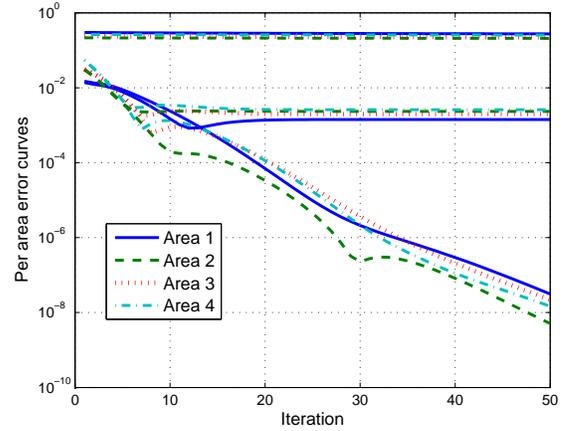}
\caption{Per area error curves $e_{k,c}^t$'s (bottom) and $e_{k,o}^t$'s (middle) for the decentralized LSE of the IEEE 14-bus system of Fig.~\ref{fig:ieee14}. The almost flat curves (top) show the corresponding $e_{k,c}^t$ error curves for the algorithm of \cite{XieChoiKar11}.}
\label{fig:14c+o+xie}
\end{figure}

\subsection{Testing the Decentralized LSE}\label{subsec:simulations:LS}
States are initialized here to the flat profile. Even though iterations \eqref{eq:Csteps} are guaranteed to converge to the optimal value of \eqref{eq:global_problem} for any $c>0$, the value of $c$ affects the convergence rate. After scaling the data to obey the model in \eqref{eq:model}, $c$ is empirically set to $10^{4}$. Two performance metrics are adopted: the per area error to the centralized solution of \eqref{eq:global_problem}, denoted by $e_{k,c}^t{:=}\|\mathbf{x}_k^{(c)}{-}\mathbf{x}_k^t\|_2/N_k$, and the per area error to the true underlying state defined as $e_{k,o}^t{:=}\|\mathbf{x}_k{-}\mathbf{x}_k^t\|_2/N_k$.

\begin{figure}[t]
\centering
\includegraphics[width=0.82\linewidth]{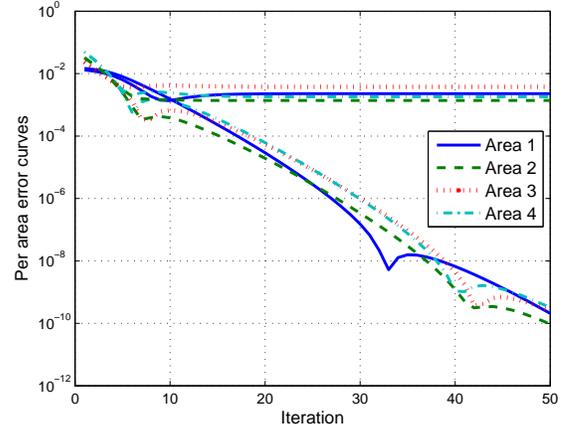}
\caption{Per area error curves $e_{k,c}^t$'s (bottom) and $e_{k,o}^t$'s (top) for the decentralized LSE of the IEEE 14-bus system without local observability.}
\label{fig:14rc+o}
\end{figure}

\begin{figure}[t]
\centering
\includegraphics[width=0.82\linewidth]{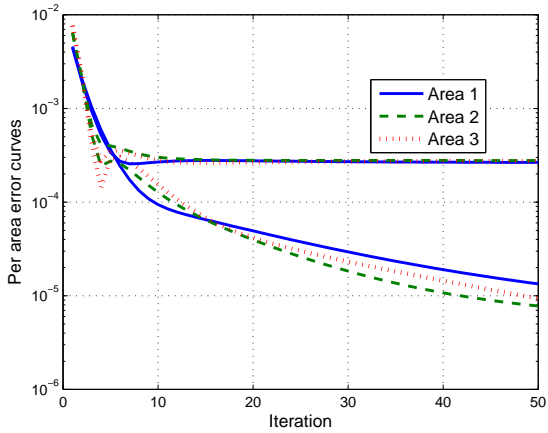}
\caption{Per area error curves $e_{k,c}^t$'s (bottom) and $e_{k,o}^t$'s (top) for the decentralized LSE of the IEEE 118-bus system of \cite[Fig.~4]{MinAbur05}.}
\label{fig:118c+o}
\end{figure}

Fig.~\ref{fig:14c+o+xie} depicts the $e_{k,c}^t$ and the $e_{k,o}^t$ curves obtained for the IEEE 14-bus network. The almost flat $e_{k,c}^t$ curves shown at the top of the figure correspond to the decentralized algorithm of \cite{XieChoiKar11} whose step sizes have been optimized. Based on the $e_{k,c}^t$ curves, the algorithm of \eqref{eq:Csteps} converges to the centralized solution. Interestingly though, as indicated by the $e_{k,o}^t$ curves, accuracy of around $10^{-3}$ dictated by the measurements is reached in 10-15 iterations. On the other hand, the algorithm of \cite{XieChoiKar11} attains the same accuracy after around 10,000 iterations. Being a first-order method, the algorithm in \cite{XieChoiKar11} incurs per iteration complexity much smaller than \eqref{eq:Csteps}, yet it does not fully exploit the capabilities of local PSSE solvers. Moreover, the high number of iterations required translates to increased inter-area communication overhead.

To evaluate the new algorithm in scenarios where local observability does not hold, the electric current measurement on line $(6,11)$ is removed from the IEEE 14-bus measurement set (cf. Fig.~\ref{fig:ieee14}). Since the only measurement directly related to bus 11 is the current measurement on line $(10,11)$ and that is collected by control area 4, area 3 is locally unobservable. The error curves obtained and plotted in Fig.~\ref{fig:14rc+o} verify that the developed method does not require local observability.

Switching to the IEEE 118-bus benchmark, similar results are observed. As evidenced by the $e_{k,c}^t$ and $e_{k,o}^t$ curves plotted in Fig.~\ref{fig:118c+o}, the decentralized solution attains the desired statistical accuracy within only 5-10 iterations.

\subsection{Testing the Decentralized Robust Estimator}\label{subsec:simulations:Huber}
The centralized versions of bad data analysis methods are compared first. The IEEE 14-bus grid of Fig.~\ref{fig:ieee14} is considered under the following four scenarios. \textbf{(S0)}: no bad data; \textbf{(S1)}: bad data on line $(4,7)$; \textbf{(S2)}: bad data on line current $(4,7)$ and bus voltage $5$; and \textbf{(S3)}: bad data on line current $(4,7)$, tie line current $(10,11)$, and bus voltage $5$. In all scenarios, bad data are simulated by multiplying the real and imaginary parts of the actual measurement by $1.2$. The performance metric here is the $\ell_2$-norm between the true state and the PSSE, which is averaged over 1,000 Monte Carlo runs.

Table~\ref{tbl:outliers} lists the results obtained by the four centralized algorithms tested: (a) an ideal yet practically infeasible genie-aided LSE (GS-LSE), which ignores the corrupted measurements; (b) the regular LSE; (c) the LNR test-based (LNRT) estimator with the test threshold set to 3.0 \cite{AburExpositoBook}; and (d) Huber's estimator of \eqref{eq:ell_1} with $\lambda=1.34$ and $\mathcal{X}=\mathbb{R}^N$. For (S0)-(S1), the estimators perform almost similarly. The few corrupted measurements in (S2)-(S3) can deteriorate LSE's performance, while Huber's estimator performs slightly better than LNRT. Computationally, Huber's estimator is implemented using iterations \eqref{eq:x_update}-\eqref{eq:o_update} for the interconnection-wide vectors $\mathbf{x}$ and $\mathbf{o}$ with $c=0$. The algorithm is terminated when the $\ell_2$-norm between the two last state iterates becomes less than $10^{-4}$. On the average and for all scenarios (S0)-(S3), Huber's estimator converges in 6-12 iterations and within 1.3~msec, while LNRT requires 2.6 recalculations of \eqref{eq:LNR} in 1.5~msec. The computing times were also measured for the IEEE 118-bus grid without corrupted data. Interestingly, the average time on the IEEE 118-bus grid without corrupted data are 3.2~msec and 81~msec, respectively. Of course, efficient updates for LNRT can be devised, but their decentralized implementation is not obvious.

\begin{table}
\renewcommand{\arraystretch}{1.2}
\caption{Mean-Square Estimation Error in the Presence of Bad Data}
\vspace*{-1em}
\centering
\begin{tabular}{|c|r|r|r|r|r|}
\hline
\textbf{Method} & GA-LSE & LSE & LNRT & Huber's \\
\hline
(S0)			& $0.0278$	& $0.0278$	& $0.0286$	& $0.0281$\\
(S1)			& $0.0313$	& $0.0318$	& $0.0331$	& $0.0322$\\
(S2)			& $0.0336$	& $0.1431$	& $0.0404$	& $0.0390$\\
(S3)			& $0.0367$	& $0.1434$	& $0.0407$	& $0.0390$\\
\hline
\end{tabular}
\label{tbl:outliers}
\end{table}

\begin{figure}
\centering
\includegraphics[width=0.82\linewidth]{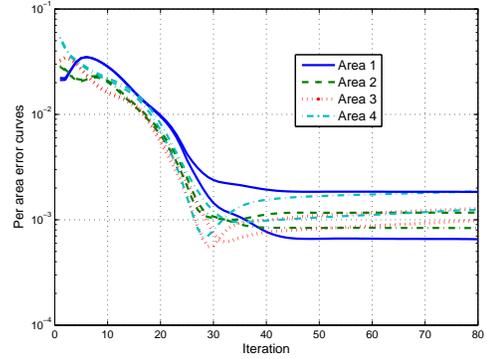}
\caption{Per area error curves $e_{k,c}^t$'s (bottom) and $e_{k,o}^t$'s (top) for the D-RPSSE algorithm on the IEEE 14-bus benchmark under (S3).}
\label{fig:rob14}
\end{figure}

\begin{figure}[t]
\centering
\includegraphics[width=0.82\linewidth]{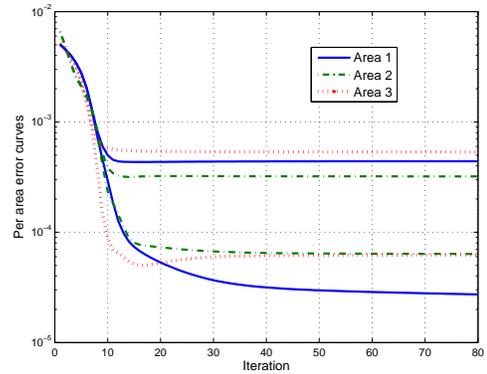}
\caption{Per area error curves $e_{k,c}^t$'s (bottom) and $e_{k,o}^t$'s (top) for the D-RPSSE algorithm on the IEEE 118-bus benchmark having 10\% of the measurements corrupted.}
\label{fig:rob118}
\end{figure}

\textcolor{black}{Focusing on decentralized implementation, the D-RPSSE algorithm (cf. Alg.~\ref{alg:DRPSSE}) is considered next. Technically, the $\hat{\mathbf{x}}$ minimizing \eqref{eq:ell_1} may not be unique. However, a sufficient condition for its uniqueness is provided in \cite[Th.~3.6]{LiSwe98}, and this condition was satisfied in all the remaining tests. D-RPSSE was first tested on the IEEE 14-bus grid under scenario (S3). The associated $e_{k,c}^t$ and $e_{k,o}^t$ curves are depicted in Fig.~\ref{fig:rob14}. The decentralized iterates approach the underlying state at an accuracy of $10^{-3}$ in 30 iterations in 12.1~msec. In comparison, the respective time for the LSE was 5~msec. Finally, for the IEEE 118-bus system, 10\% of the measurements are corrupted in the way described earlier. The corresponding error curves are plotted in Fig.~\ref{fig:rob118}. The time needed to achieve a $10^{-3}-10^{-4}$ accuracy (10 iterations) is 12.1~msec versus 3.7~msec for the related LSE.}

\color{black}

The decentralized algorithms were finally tested on a larger power network: a 4,200-bus power grid that was generated using the IEEE 14- and 300-bus grids as follows. Each one of the 300 buses of the latter is assumed to be a different area, and is replaced by a copy of the IEEE 14-bus grid. Additionally, every branch of the IEEE 300-bus grid is now an inter-area line whose terminal buses are randomly selected from the two incident to this line areas. Measurements, bad data, and $c$ are selected as in the tests for the IEEE 118-bus grid. 

\begin{figure}[t]
\centering
\includegraphics[width=0.82\linewidth]{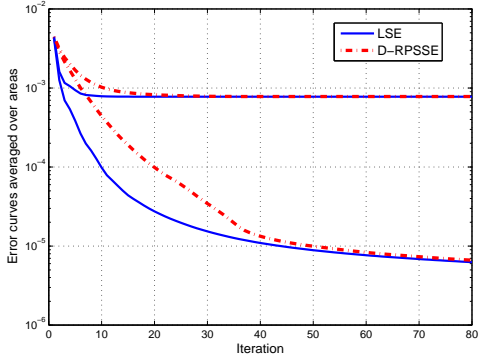}
\caption{Average error curves $\sum_{k=1}^{300}e_{k,c}^t/300$ (bottom) and $\sum_{k=1}^{300}e_{k,o}^t/300$ (top) for the LSE and D-RPSSE algorithms on a 4,200-bus grid.}
\label{fig:14x300}
\end{figure}

Fig.~\ref{fig:14x300} shows the corresponding error curves that are now averaged over the 300 areas. Given bad data-free measurements, the decentralized LSE approaches the underlying state at an accuracy of $10^{-3}$ in approximately 10 iterations or 6.2~msec; the centralized LSE finished in 93.4~msec. For a 10\% of the measurements being bad, D-RPSSE yields an accuracy of $10^{-3}$ in less than 20 ADMM iterations or 15.2~msec; the centralized robust estimator needed 193.5~msec. The convergence to their unique centralized counterparts is illustrated by the bottom curves. These tests corroborate that (i) the decentralized algorithms are basically insensitive to the variation of $c$; and (ii) the convergence time for both the decentralized LSE and D-RPSSE scale favorably with the network size. It is worth mentioning that the reverse topology where each of the 14 nodes of the IEEE 14-bus grid is replaced by a IEEE 300-bus system was tested too. Under this second architecture, the algorithms converged even faster due to the looser area coupling.

\color{black}

\section{Conclusions}\label{sec:conclusions}
Distributed and robust state estimators have been treated here under a systematic manner. The proposed algorithms waive local observability requirements and maintain backward compatibility. With a few minimal data exchanges between neighboring areas, local control centers can acquire highly accurate estimates for the part of the interconnection they are responsible for, and simultaneously identify (un)intentionally corrupted data. \textcolor{black}{The novel framework accommodates several important modifications of the PSSE problem, such as constraints (e.g., zero-injection buses, operational limits) and different MLEs. Exciting issues that emerge for future research include this work's application to generalized state estimation, its applicability to non-convex problems, and re-weighted versions of \eqref{eq:ell_1} \cite{VKGG11}.}

\appendix\label{sec:appendix}
A useful lemma is shown first.

\begin{lemma}\label{le:zero_sum}
For every pair of adjacent areas $k$ and $l$, the Lagrange multipliers updated by \eqref{eq:Asteps3} satisfy $\mathbf{v}_{k,l}^t+\mathbf{v}_{l,k}^t=\mathbf{0}$ per iteration $r>0$.
\end{lemma}

\begin{IEEEproof} 
Note that step \eqref{eq:Asteps2} decouples over the $\mathbf{x}_{kl}$'s, while the minimizers can be shown to be 
\begin{align}\label{eq:lemma_xkl_1}
\mathbf{x}_{kl}^{t+1}:=\left(\frac{\mathbf{x}_k^{t+1}[l]+\mathbf{x}_l^{t+1}[k]}{2}\right) + \left(\frac{\mathbf{v}_{k,l}^{t}+\mathbf{v}_{l,k}^{t}}{2c}\right).
\end{align}

Next, consider the updates of $\mathbf{v}_{k,l}$ and $\mathbf{v}_{l,k}$ according to step \eqref{eq:Asteps3}. Adding the two updates by parts and solving for the common term $\mathbf{x}_{kl}^{t+1}$, yields
\begin{align*}
\mathbf{x}_{kl}^{t+1}=&
\frac{\mathbf{v}_{k,l}^t+\mathbf{v}_{l,k}^t}{2c}
-\frac{\mathbf{v}_{k,l}^{t+1}+\mathbf{v}_{l,k}^{t+1}}{2c}+ \frac{\mathbf{x}_k^{t+1}[l]+\mathbf{x}_l^{t+1}[k]}{2}.
\end{align*}
By equating the right-hand sides of \eqref{eq:lemma_xkl_1} and the last equation, the claim of the lemma follows readily.
\end{IEEEproof}

\begin{IEEEproof}[Proof of Proposition \ref{pr:iterations}]
The optimization in \eqref{eq:Asteps1} is separable across areas. Upon completing the squares, the optimization task for area $k$ during step \eqref{eq:Asteps1} becomes
\begin{align}\label{eq:Asteps1b}
\min_{\mathbf{x}_k}f_k(\mathbf{x}_k) + \frac{c}{2}\sum_{l\in \mathcal{N}_k} \left\|\mathbf{x}_k[l]-\left(\mathbf{x}_{kl}^t-\frac{\mathbf{v}_{k,l}^t}{c}\right)\right\|_2^2.
\end{align}
Apparently, the $\ell_2$-norms in \eqref{eq:Asteps1b} decouple over the entries of the vectors involved. However, a single entry of $\mathbf{x}_k$, say $x_k(i)$, may be shared not only between areas $k$ and $l$, but rather among area $k$ and all the areas in $\mathcal{N}_k^i$. If $x_{kl}[i]$ $\left(v_{k,l}[i]\right)$ denotes the entry of $\mathbf{x}_{kl}$ $\left(\mathbf{v}_{k,l}\right)$ corresponding to $x_{k}(i)$, the optimization in \eqref{eq:Asteps1b} can be expressed as
\begin{align}\label{eq:Asteps1c}
\min_{\mathbf{x}_k}f_k(\mathbf{x}_k) + \frac{c}{2} \sum_{\substack{i\in \mathcal{N}_k\\\mathcal{N}_k^i\neq \emptyset}} |\mathcal{N}_k^i|\left(x_k(i)-p_k^{t+1}(i)\right)^2
\end{align}
where for all $k$, and $i=1,\ldots,N_k$ with $\mathcal{N}_k^i\neq \emptyset$, define
\begin{align}\label{eq:p}
p_k^{t+1}(i):=\frac{1}{|\mathcal{N}_k^i|}\sum_{l\in \mathcal{N}_k^i}\left(x_{kl}^t[i]-\frac{v_{k,l}^t[i]}{c}\right).
\end{align}

By Lemma \ref{le:zero_sum}, step \eqref{eq:Asteps2} simplifies to $\mathbf{x}_{kl}^{t+1} = 0.5 \left(\mathbf{x}_k^{t+1}[l]+\mathbf{x}_l^{t+1}[k]\right)$. In other words, the auxiliary variable $\mathbf{x}_{kl}$ is the average of the shared state variables across areas $k$ and $l$ per iteration. By eliminating the auxiliary variables $\mathbf{x}_{kl}$ from the updates of \eqref{eq:p} and \eqref{eq:Asteps3}, step \eqref{eq:Asteps2} can be dropped. Hence, one arrives at the iterates 

\small
\begin{subequations}\label{eq:Bsteps}
\begin{align}
&\mathbf{x}_k^{t+1}:=\arg\min_{\mathbf{x}_k}f_k(\mathbf{x}_k) + \frac{c}{2} \sum_{\substack{i\in \mathcal{N}_k\\ \mathcal{N}_k^i\neq \emptyset}}|\mathcal{N}_k^i|\left(x_k(i){-}p_k^t(i)\right)^2\label{eq:Bsteps1}\\
&\mathbf{v}_{k,l}^{t+1}:=\mathbf{v}_{k,l}^t + c\left(\frac{\mathbf{x}_k^{t+1}[l]-\mathbf{x}_l^{t+1}[k]}{2}\right)\label{eq:Bsteps2}\\
&p_k^{t+1}(i):=\frac{1}{2}\left(x_k^{t+1}(i)+ \frac{1}{|\mathcal{N}_k^i|}\sum_{l\in \mathcal{N}_k^i}x_{l}^{t+1}[i]\right){-} \frac{1}{|\mathcal{N}_{k}^i|}\sum_{l\in \mathcal{N}_k^i}\frac{v_{k,l}^{t+1}[i]}{c}.
\label{eq:Bsteps3}
\end{align}
\end{subequations}
\normalsize

To further simplify the iterations, define the average of the shared variable $x_k(i)$'s copies over $\mathcal{N}_k^i$ as the $s_k^t(i)$ in \eqref{eq:Csteps2}. Define also the average of the weighted Lagrange multipliers $u_k^t(i):=\sum_{l\mathcal{N}_k^i}v_{k,l}^t[i]/(c|\mathcal{N}_k^i|)$. Then, \eqref{eq:Bsteps3} can be written as
\begin{align}\label{eq:p_non_recursive_update}
p_k^{t+1}(i):=\frac{1}{2}\left(x_k^{t+1}(i) + s_k^{t+1}(i)\right) - u_k^{t+1}(i).
\end{align}
With $\{\mathbf{v}_{k,l}\}$ initialized to zero, $\{u_k(i)\}$ can be recursively updated as $u_k^{t+1}(i):= u_k^{t}(i) + (x_k^{t+1}(i)-s_k^{t+1}(i))/2$. Hence, update \eqref{eq:p_non_recursive_update} can be alternatively performed as in \eqref{eq:Csteps3}. Collecting \eqref{eq:Bsteps1}, the definition of $\{s_k^t(i)\}$, and the recursive updates for $\{p_k^{t}(i)\}$, one readily arrives at \eqref{eq:Csteps}.
\end{IEEEproof}

%
%

\bibliographystyle{IEEEtranS}
\bibliography{IEEEabrv,power}

\begin{thebibliography}{10}
\providecommand{\url}[1]{#1}
\csname url@samestyle\endcsname
\providecommand{\newblock}{\relax}
\providecommand{\bibinfo}[2]{#2}
\providecommand{\BIBentrySTDinterwordspacing}{\spaceskip=0pt\relax}
\providecommand{\BIBentryALTinterwordstretchfactor}{4}
\providecommand{\BIBentryALTinterwordspacing}{\spaceskip=\fontdimen2\font plus
\BIBentryALTinterwordstretchfactor\fontdimen3\font minus
  \fontdimen4\font\relax}
\providecommand{\BIBforeignlanguage}[2]{{%
\expandafter\ifx\csname l@#1\endcsname\relax
\typeout{** WARNING: IEEEtranS.bst: No hyphenation pattern has been}%
\typeout{** loaded for the language `#1'. Using the pattern for}%
\typeout{** the default language instead.}%
\else
\language=\csname l@#1\endcsname
\fi
#2}}
\providecommand{\BIBdecl}{\relax}
\BIBdecl

\bibitem{AburExpositoBook}
A.~Abur and A.~Gomez-Exposito, \emph{Power System State Estimation: Theory and
  Implementation}.\hskip 1em plus 0.5em minus 0.4em\relax New York, NY: Marcel
  Dekker, 2004.

\bibitem{Boyd10}
S.~Boyd, N.~Parikh, E.~Chu, B.~Peleato, and J.~Eckstein, ``Distributed
  optimization and statistical learning via the alternating direction method of
  multipliers,'' \emph{{F}ound. {T}rends {M}ach {L}earning}, vol.~3, pp.
  1--122, 2010.

\bibitem{CelikAbur92}
M.~K. Celik and A.~Abur, ``A robust {WLAV} state estimator using
  transformations,'' \emph{{IEEE} Trans. Power Syst.}, vol.~7, no.~1, pp.
  106--113, Feb. 1992.

\bibitem{Conejo07}
A.~J. Conejo, S.~de~la Torre, and M.~Canas, ``An optimization approach to
  multiarea state estimation,'' \emph{{IEEE} Trans. Power Syst.}, vol.~22,
  no.~1, pp. 213--221, Feb. 2007.

\bibitem{cutsem83}
T.~V. Cutsem and M.~Ribbens-Pavella, ``Critical survey of hierarchical methods
  for state estimation of electric power systems,'' \emph{{IEEE} Trans. Power
  App. Syst.}, vol. 102, no.~10, pp. 3415--3424, Oct. 1983.

\bibitem{ReCeTh10}
J.~De~La~Ree, V.~Centeno, J.~Thorp, and A.~Phadke, ``Synchronized phasor
  measurement applications in power systems,'' \emph{{IEEE} Trans. Smart Grid},
  vol.~1, no.~1, pp. 20--27, Jun. 2010.

\bibitem{DuanYangScharf11}
D.~Duan, L.~Yang, and L.~L. Scharf, ``Phasor state estimation from {PMU}
  measurements with bad data,'' in \emph{Proc. {IEEE} {Workshop on Comp. Adv.
  in Multi-Sensor Adaptive Proc.}}, San Juan, Puerto Rico, Dec. 2011.

\bibitem{Ebrah00}
R.~Ebrahimian and R.~Baldick, ``State estimation distributed processing,''
  \emph{{IEEE} Trans. Power Syst.}, vol.~15, no.~4, pp. 1240--1246, Nov. 2000.

\bibitem{ElKeib92}
A.~A. El-Keib, J.~Nieplocha, H.~Singh, and D.~Maratukulam, ``A decomposed state
  estimation technique suitable for parallel processor implementation,''
  \emph{{IEEE} Trans. Power Syst.}, vol.~7, no.~3, pp. 1088--1097, Aug. 1992.

\bibitem{Falcao95}
D.~M. Falcao, F.~F. Wu, and L.~Murphy, ``Parallel and distributed state
  estimation,'' \emph{{IEEE} Trans. Power Syst.}, vol.~10, no.~2, pp. 724--730,
  May 1995.

\bibitem{ExpositoPROC11}
A.~G\'{o}mez-Exposito, A.~Abur, A.~de~la Villa~Ja\'{e}n, and
  C.~Go\'{m}ez-Quiles, ``A multilevel state estimation paradigm for smart
  grids,'' \emph{Proc. {IEEE}}, vol.~99, no.~6, pp. 952--976, Jun. 2011.

\bibitem{MASEsurvey}
A.~G\'{o}mez-Exposito, A.~de~la Villa~Ja\'{e}n, C.~Go\'{m}ez-Quiles,
  P.~Rousseaux, and T.~V. Cutsem, ``A taxonomy of multi-area state estimation
  methods,'' \emph{Electric Power Systems Research}, vol.~81, pp. 1060--1069,
  2011.

\bibitem{Iwamoto89}
S.~Iwamoto, M.~Kusano, and V.~H. Quintana, ``Hierarchical state estimation
  using a fast rectangular-coordinate method,'' \emph{{IEEE} Trans. Power
  Syst.}, vol.~4, no.~3, pp. 870--880, Aug. 1989.

\bibitem{VKGG11}
V.~Kekatos and G.~B. Giannakis, ``From sparse signals to sparse residuals for
  robust sensing,'' \emph{{IEEE} Trans. Signal Process.}, vol.~59, no.~7, pp.
  3355--3368, Jul. 2011.

\bibitem{Korres11}
G.~N. Korres, ``A distributed multiarea state estimation,'' \emph{{IEEE} Trans.
  Power Syst.}, vol.~26, no.~1, pp. 73--84, Feb. 2011.

\bibitem{Kosut11}
O.~Kosut, L.~Jia, J.~Thomas, and L.~Tong, ``Malicious data attacks on the smart
  grid,'' \emph{{IEEE} {T}rans. {S}mart {G}rid}, vol.~2, no.~4, pp. 645--658,
  Dec. 2011.

\bibitem{LiSwe98}
W.~Li and J.~J. Swetits, ``The linear {$\ell_1$ Estimator and {H}uber
  {M}-Estimator},'' \emph{SIAM J. Optim.}, vol.~8, pp. 457--475, May 1998.

\bibitem{LinLin94}
S.-Y. Lin and C.-H. Lin, ``An implementable distributed state estimator and
  distributed bad data processing schemes for electric power systems,''
  \emph{{IEEE} Trans. Power Syst.}, vol.~9, no.~3, pp. 1277--1284, Aug. 1994.

\bibitem{LiuReiNing09}
Y.~Liu, M.~K. Reiter, and P.~Ning, ``False data injection attacks against state
  estimation in electric power grids,'' in \emph{Proc. {ACM} {Conf. on Computer
  and Comm. Security}}, Chicago, IL, Nov. 2009, pp. 21--32.

\bibitem{MaMu00}
O.~L. Mangasarian and D.~R. Musicant, ``Robust linear and support vector
  regression,'' \emph{{IEEE} Trans. Pattern Anal. Mach. Intell.}, vol.~22,
  no.~9, pp. 950--955, Sep. 2000.

\bibitem{MaMaYo06}
R.~A. Maronna, R.~D. Martin, and V.~J. Yohai, \emph{Robust Statistics: Theory
  and Methods}.\hskip 1em plus 0.5em minus 0.4em\relax Wiley, 2006.

\bibitem{Mili94}
L.~Mili, M.~G. Cheniae, and P.~J. Rousseeuw, ``Robust state estimation of
  electric power systems,'' \emph{{IEEE} Trans. Circuits Syst. {I}}, vol.~41,
  no.~5, pp. 349--358, May 1994.

\bibitem{MinAbur05}
L.~Min and A.~Abur, ``Total transfer capability computation for multi-area
  power systems,'' \emph{{IEEE} Trans. Power Syst.}, vol.~21, no.~3, pp.
  1141--1147, Aug. 2006.

\bibitem{SecurityPROC12}
B.~Y. Mo, T.~H.-J. Kim, K.~Brancik, D.~Dickinson, H.~Lee \emph{et~al.},
  ``Cyber–physical security of a smart grid infrastructure,'' \emph{Proc.
  {IEEE}}, vol. 100, no.~1, pp. 195--209, Jan. 2012.

\bibitem{Mo00}
A.~Monticelli, ``Electric power system state estimation,'' \emph{Proc. {IEEE}},
  vol.~88, no.~2, pp. 262--282, Feb. 2000.

\bibitem{PhTh08}
A.~G. Phadke and J.~S. Thorp, \emph{Synchronized Phasor Measurements and Their
  Applications}.\hskip 1em plus 0.5em minus 0.4em\relax New York, NY: Springer,
  2008.

\bibitem{Schweppe70}
F.~C. Schweppe, J.~Wildes, and D.~Rom, ``Power system static state estimation:
  Parts {I, II, and III},'' \emph{{IEEE} Trans. Power App. Syst.}, vol.~89, pp.
  120--135, Jan. 1970.

\bibitem{PSTCA}
\BIBentryALTinterwordspacing
Power systems test case archive. University of Washington. [Online]. Available:
  \url{http://www.ee.washington.edu/research/pstca/}
\BIBentrySTDinterwordspacing

\bibitem{XieChoiKar11}
L.~Xie, D.-H. Choi, and S.~Kar, ``Cooperative distributed state estimation:
  Local observability relaxed,'' in \emph{Proc. {IEEE} {PES} {G}eneral
  {M}eeting}, Detroit, MI, Jul. 2011.

\bibitem{XuWaTang11}
\BIBentryALTinterwordspacing
W.~Xu, M.~Wang, and A.~Tang, ``Sparse recovery from nonlinear measurements with
  applications in bad data detection for power networks,'' submitted, Dec.
  2011. [Online]. Available: \url{http://arxiv.org/abs/1112.6234}
\BIBentrySTDinterwordspacing

\bibitem{ZhaoAbur05}
L.~Zhao and A.~Abur, ``Multiarea state estimation using synchronized phasor
  measurements,'' \emph{{IEEE} Trans. Power Syst.}, vol.~20, no.~2, pp.
  611--617, May 2005.

\bibitem{Zhu09}
H.~Zhu, G.~B. Giannakis, and A.~Cano, ``Distributed in-network decoding,''
  \emph{{IEEE} Trans. Signal Process.}, vol.~57, no.~10, pp. 3970--3983, Oct.
  2009.

\bibitem{MATPOWER}
R.~D. Zimmerman, C.~E. Murillo-Sanchez, and R.~J. Thomas, ``{MATPOWER}:
  steady-state operations, planning and analysis tools for power systems
  research and education,'' \emph{{IEEE} Trans. Power Syst.}, vol.~26, no.~1,
  pp. 12--19, Feb. 2011.

\end{thebibliography}

\begin{IEEEbiography}[{\includegraphics[width=1in,height=1.25in,clip,keepaspectratio]{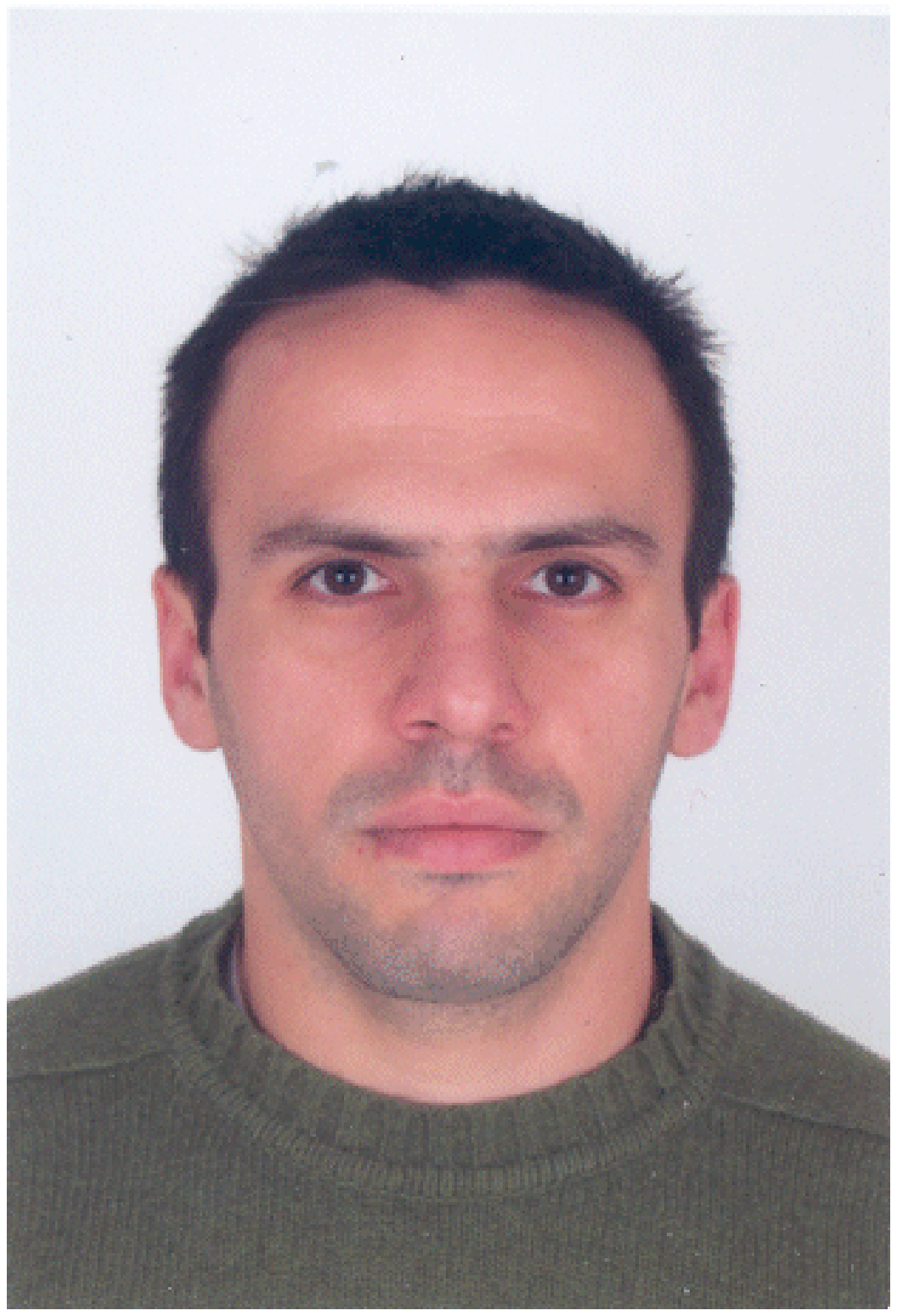}}]{Vassilis Kekatos} (M'10) received the Diploma, MSc., and PhD in Computer Engr. \& Informatics Dept. from the University of Patras, Greece, in 2001, 2003, and 2007, respectively. Since 2009, he has been a Marie Curie Fellow, and he is currently a post doctoral associate with the Dept. of Electrical and Computer Engr. at the University of Minnesota, and the Computer Engnr. \& Informatics Dept. at the University of Patras, Greece. His research interests lie in the areas of statistical signal processing with emphasis on the power grid, compressive sampling, and wireless communications.
\end{IEEEbiography}

\begin{IEEEbiography}[{\includegraphics[width=1in,height=1.25in,clip,keepaspectratio]{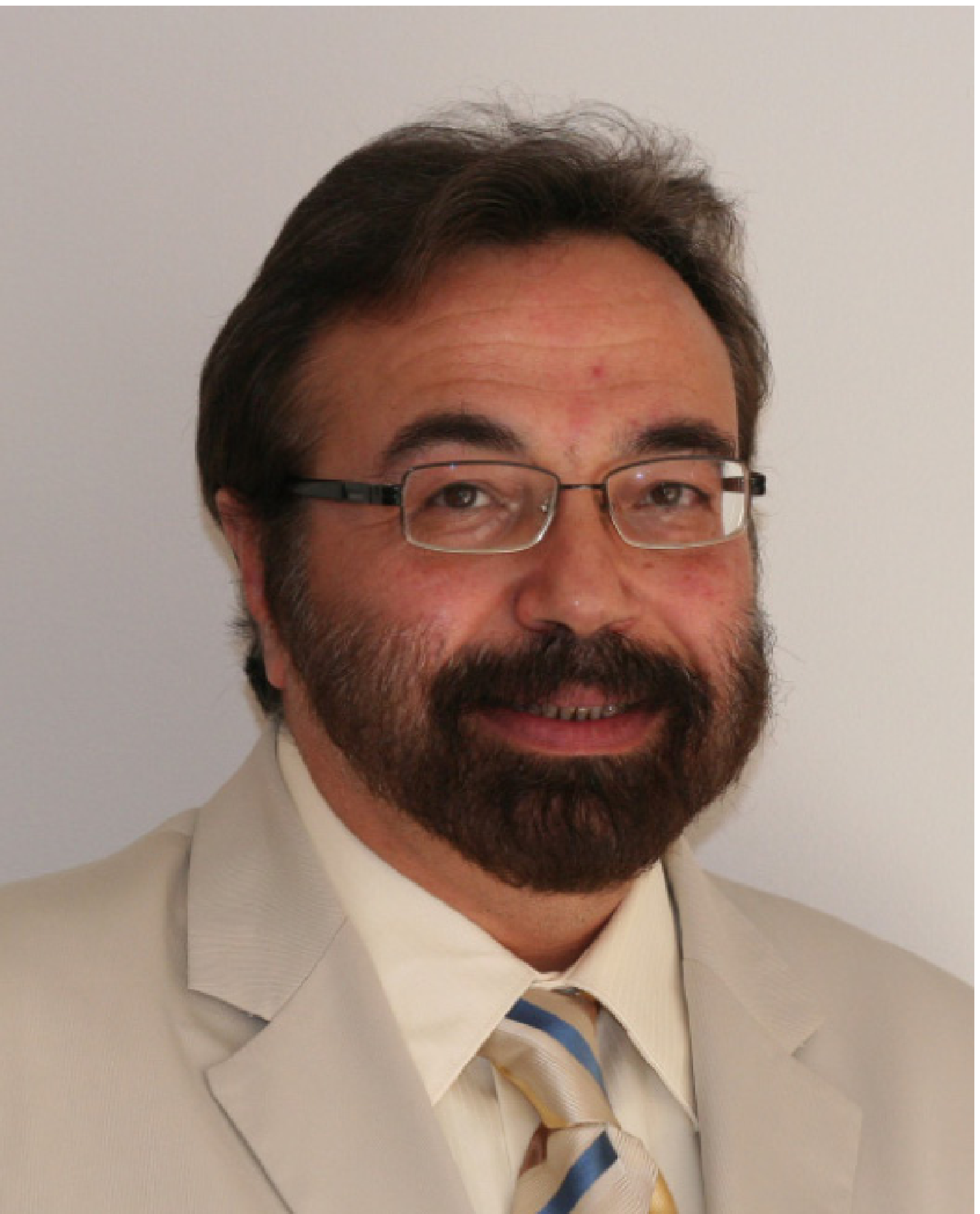}}] {Georgios B. Giannakis} (F'97) received his Diploma in Electrical Engr. from the Ntl. Tech. Univ. of Athens, Greece, 1981. From 1982 to 1986 he was with the Univ. of Southern California (USC), where he received his MSc. in Electrical Engineering, 1983, MSc. in Mathematics, 1986, and Ph.D. in Electrical Engr., 1986. Since 1999 he has been a professor with the Univ. of Minnesota, where he now holds an ADC Chair in Wireless Telecommunications in the ECE Department, and serves as director of the Digital Technology Center.

His general interests span the areas of communications, networking and statistical signal processing - subjects on which he has published more than 325 journal papers, 525 conference papers, 20 book chapters, two edited books and two research monographs. Current research focuses on compressive sensing, cognitive radios, network coding, cross-layer designs, wireless sensors, social and power grid networks. He is the (co-)inventor of 21 patents issued, and the (co-)recipient of eight paper awards from the IEEE Signal Processing (SP) and Communications Societies, including the G. Marconi Prize Paper Award in Wireless Communications. He also received Technical Achievement Awards from the SP Society (2000), from EURASIP (2005), a Young Faculty Teaching Award, and the G. W. Taylor Award for Distinguished Research from the University of Minnesota. He is a Fellow of EURASIP, and has served the IEEE in a number of posts, including that of a Distinguished Lecturer for the IEEE-SP Society.
\end{IEEEbiography}

\end{document}